\PassOptionsToPackage{dvipsnames}{xcolor}
\documentclass[letterpaper, 10 pt, conference]{ieeeconf}

\usepackage[utf8]{inputenc} 
\usepackage[T1]{fontenc}    
\usepackage{url}            
\usepackage{booktabs}       
\usepackage{nicefrac}       
\usepackage{microtype}      
\usepackage{xcolor}         
\usepackage{colortbl}
\usepackage{graphicx}
\usepackage{amsmath,amsfonts,amssymb}
\usepackage{color}
\usepackage{caption}
\usepackage{subcaption}
\usepackage{xspace}
\usepackage{array}
\usepackage{cite}
\usepackage{hyperref}
\usepackage{multirow}
\usepackage{todonotes}
\usepackage{pifont}  

\makeatletter
\DeclareRobustCommand\onedot{\futurelet\@let@token\@onedot}
\def\@onedot{\ifx\@let@token.\else.\null\fi\xspace}

\makeatother

\newcommand{\boldparagraph}[1]{\vspace{0.0cm}\noindent{\bf #1.} }

\definecolor{darkgreen}{rgb}{0,0.7,0}
\definecolor{lgray}{rgb}{0.83,0.83,0.83}
\definecolor{ellisdarkgreen}{rgb}{0.47,0.76,0.24}

\definecolor{ellisred}{rgb}{0.87,0.44,0.38} 
\definecolor{ellisgreen}{rgb}{0.69,0.90,0.52} 
\definecolor{elliscyan}{rgb}{0.29,0.77,0.74} 
\definecolor{ellisorange}{rgb}{0.89,0.55,0.28} 
\definecolor{ellisblue}{rgb}{0.41,0.61,0.86} 
\definecolor{ellispurple}{rgb}{0.44,0.19,0.63} 

\newcommand{\ellisred}[1]{\noindent{\color{ellisred}{#1}}}
\newcommand{\ellisgreen}[1]{\noindent{\color{ellisgreen}{#1}}}

\newcommand{\ellisorange}[1]{\noindent{\color{ellisorange}{#1}}}
\newcommand{\ellisblue}[1]{\noindent{\color{ellisblue}{#1}}}
\newcommand{\ellispurple}[1]{\noindent{\color{ellispurple}{#1}}}

\newcommand{\nuplan}{nuPlan\xspace}
\newcommand{\nuscenes}{nuScenes\xspace}
\newcommand{\sota}{state-of-the-art\xspace}
\newcommand{\interplan}{\textit{interPlan}\xspace}

\newcommand{\red}[1]{\textcolor{red}{#1}}
\newcommand{\forrestgreen}[1]{\textcolor{ForestGreen}{#1}}
\newcommand{\cmark}{\forrestgreen{\ding{51}}}  
\newcommand{\xmark}{\red{\ding{55}}}  

\newlength{\quarterimgwidth}
\title{Can Vehicle Motion Planning Generalize to Realistic Long-tail Scenarios?}
\author{Marcel Hallgarten$^{1,2}$ \quad Julian Zapata$^{2,3}$ \quad Martin Stoll$^{2}$ \quad
Katrin Renz$^{1,4}$ \quad Andreas Zell$^{1}$\\
 $^1$University of Tübingen \quad $^2$Robert Bosch GmbH \quad $^3$University of Duisburg-Essen \quad $^4$Tübingen AI Center\\
 {\tt\small \href{https://github.com/mh0797/interPlan}{{\color{magenta}{https://github.com/mh0797/interPlan}}}}
}
\date{January 2024}

\begin{document}

\maketitle

\begin{abstract}
Real-world autonomous driving systems must make safe decisions in the face of rare and diverse traffic scenarios.
Current state-of-the-art planners are mostly evaluated on real-world datasets like \nuscenes (open-loop) or \nuplan (closed-loop).
In particular \nuplan seems to be an expressive evaluation method since it is based on real-world data and closed-loop, yet it mostly covers basic driving scenarios.
This makes it difficult to judge a planner's capabilities to generalize to rarely-seen situations.
Therefore, we propose a novel closed-loop benchmark \interplan containing several edge cases and challenging driving scenarios.
We assess existing state-of-the-art planners on our benchmark and show that neither rule-based nor learning-based planners can safely navigate the \interplan scenarios.

A recently evolving direction is the usage of foundation models like large language models (LLM) to handle generalization.
We evaluate an LLM-only planner and introduce a novel hybrid planner that combines an LLM-based behavior planner with a rule-based motion planner that achieves state-of-the-art performance on our benchmark.
\end{abstract}

\section{Introduction}
Generalization to previously unseen driving scenarios is crucial to achieving autonomy and motivates research on learning-based vehicle motion planning methods.
Yet, simple rule-based planners outperform learning-based methods in the recently proposed closed-loop \nuplan benchmark~\cite{caesar2021nuplan}.
Even more surprisingly, the rule-based method PDM-Closed~\cite{Dauner2023CORL} achieves a nearly perfect score, suggesting that it is capable of tackling the enormous challenge of real-world driving.
In this work, we challenge this conception.
The fact that PDM-Closed~\cite{Dauner2023CORL} is a reactive centerline planner, unable to do lane changes or interact in a targeted way with surrounding traffic agents, underlines that the \nuplan benchmark mostly covers basic driving scenarios.
While solving these is a fundamental requirement for motion planning methods, it falls short of proving that they generalize to rare scenarios from the long-tailed distribution of potential real-world scenarios. 

\boldparagraph{The interPlan benchmark}
During complex traffic maneuvers there is often mutual influence between traffic participants.
Mastering these bi-directional interactions is crucial for smooth progress,
ultimately making it a fundamental enabler for autonomy~\cite{hagedorn2023rethinking, rhinehart2021contingencies}.
Therefore, we propose a novel benchmark to evaluate vehicle motion planning methods in highly interactive and rare scenarios.
We take \nuplan~\cite{caesar2021nuplan} scenarios as a starting point for our benchmark and augment them with additional agents, obstacles, or alternative navigation goals, creating realistic long-tail scenarios.
In total, our benchmark comprises 80 scenarios covering the following situations: Nudging around parked vehicles, overtaking obstacles, passing a construction zone, passing an accident site, facing jaywalkers, as well as lane changes in low, medium, and high traffic density.
A comparative overview is shown in Tab.~\ref{tab:comparison}.

\boldparagraph{State-of-the-art methods do not generalize to long-tail scenarios}
We conduct experiments with an exhaustive set of state-of-the-art planning methods and reveal critical shortcomings in their ability to generalize to difficult unseen scenarios.
Rule-based vehicle motion planning methods rely on predefined rules and behaviors to navigate the vehicle.
They fail to handle complex and uncommon driving scenarios not covered by the limited rules.
On the other hand, learning-based methods are impaired by a lack of robustness and thus cannot deliver the promise of better generalization.

\boldparagraph{LLM-based planning}
Foundation models, especially Large Language Models (LLMs), have shown outstanding world-understanding and generalization ability.
This makes them tailored to analyze complex driving environments and make informed and well-founded decisions.
Therefore, we propose LLM-based planning baselines for our novel benchmark and analyze their capabilities in closed-loop simulation.


\begin{table}[t]
    \centering
    \resizebox{\columnwidth}{!}{
    \begin{tabular}{l c c c}
        \toprule
         & nuPlan~\cite{caesar2021nuplan} & Carla~\cite{dosovitskiy2017carla} & interPlan (ours)\\
         \midrule
         Task & \forrestgreen{Planning} & \ellisorange{E2E Driving} & \forrestgreen{Planning} \\
         Based on real-world scenarios & \cmark & \xmark & \cmark \\
         Diverse traffic agents & \xmark & \xmark & \cmark \\
         Long-tail scenarios & \xmark & \cmark & \cmark \\
         Interactive lane-changes &\xmark & \cmark & \cmark \\
         Scenario-generation interface & \xmark & \cmark & \cmark \\
         LLM interface & \xmark & \xmark & \cmark \\
         \bottomrule
    \end{tabular}
    }
    \caption{Comparative summary of planning benchmarks}
    \label{tab:comparison}
    \vspace{-0.6cm}
\end{table}

\boldparagraph{Contributions}
We summarize our contributions as follows:
\begin{itemize}
    \item We propose the \interplan benchmark, a publicly available closed-loop driving benchmark focused on highly interactive and difficult scenarios.
    We evaluate and analyze a comprehensive set of \sota methods.
    \item We demonstrate that even though some methods achieve excellent results in common driving scenarios, they fail in complex long-tail situations, e.g., passing accident sites.
    \item Fueled by the rising interest in motion planning based on LLMs, we implement GPT-Driver~\cite{mao2023gpt} as a baseline and challenge its abilities.
    \item We propose a novel two-stage planner that combines an LLM-based behavior planner with a downstream rule-based motion planner.
    It led to a new state-of-the-art on the novel benchmark and serves as a strong baseline for future research.
\end{itemize}
\section{Related Work}

\boldparagraph{Closed-loop vehicle motion planning}
Training a learning-based planner through imitation learning is a straightforward and widely adopted approach~\cite{chitta2021neat,chitta2022transfuser,zeng2019end,rhinehart2019precog,codevilla2019exploring,codevilla2018end} pioneered by ALVINN~\cite{pomerleau1988alvinn}.
Rule-based methods, on the other hand, employ a hand-designed decision-making framework with interpretable rules~\cite{thrun2006stanley, bacha2008odin, leonard2008perception, urmson2008autonomous, chen2015deepdriving, sauer2018conditional, fan2018baidu}.
For instance, the Intelligent Driver Model (IDM)~\cite{treiber2000congested} always follows a lane-centerline while maintaining a safe distance to the leading vehicle.
Two predominant methods exist in evaluating vehicle motion planners.
Open-loop ego forecasting compares the planned trajectory to an expert's ground-truth trajectory using distance-based metrics.
In contrast, closed-loop evaluation involves simulating the planned trajectory with realistic environmental feedback, such as vehicle dynamics and reactions of surrounding traffic agents.
To assess the interactive behavior in complex and interactive scenarios, closed-loop evaluation is indispensable, particularly as open-loop ego-forecasting evaluation was shown to be misaligned with closed-loop driving~\cite{Dauner2023CORL,codevilla2018offline}.
The most widely adopted driving simulators are CARLA~\cite{dosovitskiy2017carla} and \nuplan~\cite{caesar2021nuplan}.
The simplistic CARLA~\cite{dosovitskiy2017carla} simulator uses synthetic data, which is not guaranteed to produce realistic driving situations.
On the other hand, \nuplan~\cite{caesar2021nuplan} is based on real-world data, which comes at the cost of undersampling rare and critical scenarios, since collecting them is prohibitive due to the cost and danger involved.
For instance, obtaining recordings of an emergency break caused by a child running on the street is dangerous and unethical.
While the Carla Leaderboard fueled research for end-to-end methods that directly process sensor information~\cite{chen2022learning, chitta2022transfuser, zeng2019end, jaeger2023hidden, shao2023safety, wu2022trajectory, Shao_2023_CVPR}, the \nuplan competition focused on modular planners that use outputs of upstream perception modules (e.g., bounding boxes, maps) as inputs~\cite{Scheel2021CORL, renz2022plant, Hallgarten2022ITSC, huang2023gameformer, Huang2023ICRA, cheng2023rethinking}.
In this work, we make use of the \nuplan closed-loop simulation to assess planners independently from perception noise.
In order to assess generalization capabilities to rare driving situations, our proposed \interplan benchmark is made up of augmented real-world scenarios.

\boldparagraph{Benchmarking in complex interactive scenarios}
Autonomous driving systems need to meet high safety standards.
In particular, they must be extraordinarily robust when confronted with unseen scenarios.
Various approaches exist to generate safety-critical scenarios for the synthetic Carla simulator~\cite{dosovitskiy2017carla,ding2020learning, ding2021multimodal, wang2021advsim}.
These methods focus on robustness in the face of rule-breaking or assertive agents and behaviors that mislead the ego-vehicle (EV) into questionable decisions.
Other approaches include gradient-based optimization~\cite{hanselmann2022king, rempe2022generating} and prior-knowledge based scenario construction~\cite{bagschik2018ontology, mcduff2022causalcity, menzel2018scenarios}.
Similarly, we augment driving scenarios to assess the interactive behavior of planning methods, facing a variety of traffic conditions ranging from sparse conservative traffic to high-density assertive traffic.
Moreover, we also test scenarios that are not safety-critical but require profound reasoning and robust generalization, such as passing an accident site or a construction zone.
In contrast to previous work, we base our scenario generation on real-world scenarios from the widely adopted \nuplan dataset and simulator.
We thus benefit from an unprecedented amount of typical driving scenarios for training before testing a planner's generalization to long-tail scenarios.

\boldparagraph{LLM-based vehicle motion planning}
With the success of Large Language Models (LLMs), a vital field of research emerged from applying them to vehicle motion planning.
GPT-Driver~\cite{mao2023gpt} proposes to model motion planning as a language modeling task.
By feeding outputs of a perception module~\cite{hu2023planning} into a prompt and instructing the LLM to perform chain-of-thought reasoning to ultimately output a safe motion trajectory, it achieves state-of-the-art performance on the open-loop \nuscenes benchmark~\cite{caesar2020nuscenes}.
This was extended by AgentDriver~\cite{mao2023language}, which uses another LLM to generate the task prompt.
We introduce a hybrid approach that combines the outstanding world understanding and common sense of an LLM-based behavior planner with the excellent robustness of a rule-based motion planner.
While prior work demonstrated strong reasoning capabilities of LLMs in traffic scenarios~\cite{sha2023languagempc,chen2023driving,kim2020advisable,jin2023adapt}, methods are often only evaluated in open-loop evaluation~\cite{sun2023large,mao2023gpt,hu2023planning,mao2023language, Sima2023arxiv}, in the synthetic Carla simulator~\cite{dosovitskiy2017carla,wang2023drivemlm}, or the simplistic HighwayEnv environment~\cite{leurent2018environment,fu2024drive, wang2023empowering}.
In contrast, our closed-loop \interplan benchmark is based on augmented real-world data, covers 80 complex driving scenarios, and includes a comprehensive set of baselines.
\section{Realistic Scenario Generation}
\begin{figure*}[t!]
	\centering
    \begin{subfigure}[t]{\quarterimgwidth}
		\includegraphics[width=\textwidth]{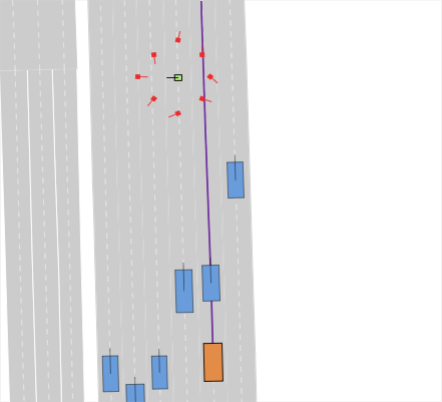}
		\caption{\footnotesize Construction site}
        \label{fig:constr}
    \end{subfigure}\hfill\begin{subfigure}[t]{\quarterimgwidth}
		\includegraphics[width=\textwidth]{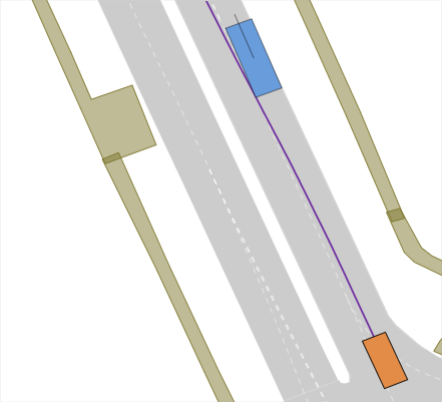}
		\caption{\footnotesize Nudge}
        \label{fig:nudge}
    \end{subfigure}\hfill\begin{subfigure}[t]{\quarterimgwidth}
		\includegraphics[width=\textwidth]{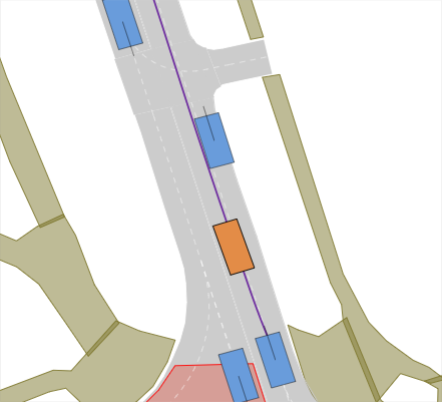}
		\caption{\footnotesize Overtake}
        \label{fig:overtake}
    \end{subfigure}\hfill\begin{subfigure}[t]{\quarterimgwidth}
		\includegraphics[width=\textwidth]{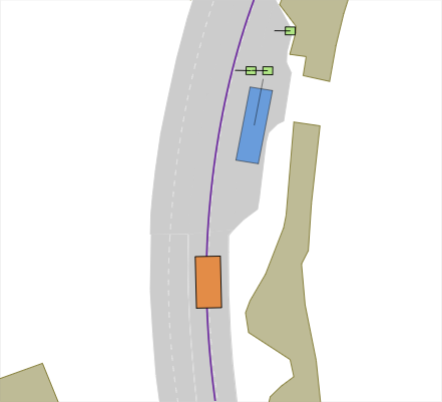}
		\caption{\footnotesize Jaywalker}
        \label{fig:jayw}
    \end{subfigure}
    \newline
    \begin{subfigure}[t]{\quarterimgwidth}
        \includegraphics[width=\textwidth]{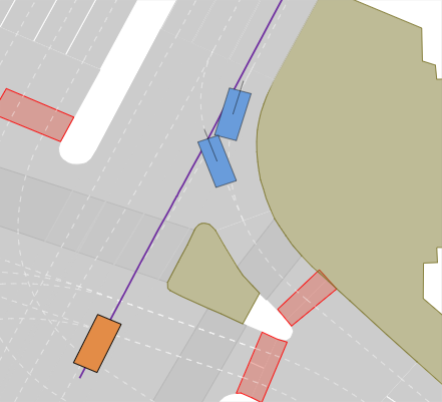}
		\caption{\footnotesize Accident site}
        \label{fig:crash}
    \end{subfigure}\hfill\begin{subfigure}[t]{\quarterimgwidth}
		\includegraphics[width=\textwidth]{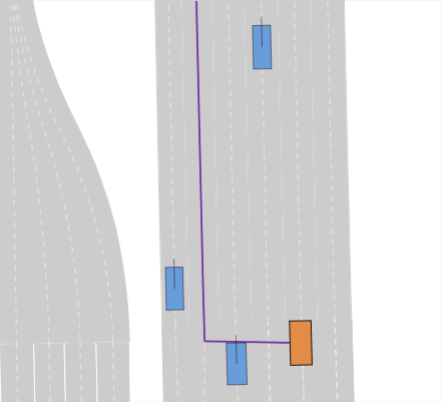}
		\caption{Lane-change (LTD)}
        \label{fig:ltd}
    \end{subfigure}\hfill\begin{subfigure}[t]{\quarterimgwidth}
		\includegraphics[width=\textwidth]{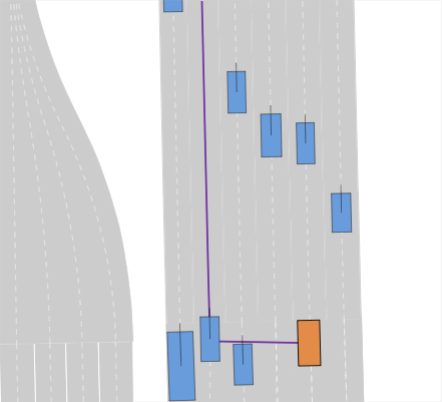}
		\caption{Lane-change (MTD)}
        \label{fig:mtd}
    \end{subfigure}\hfill\begin{subfigure}[t]{\quarterimgwidth}
		\includegraphics[width=\textwidth]{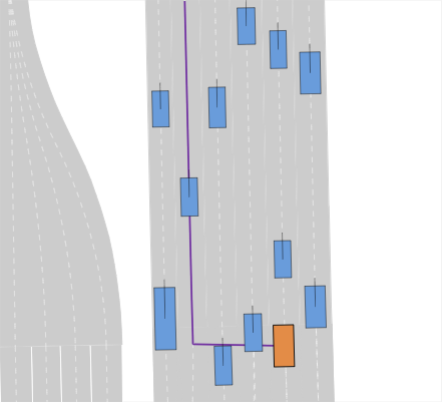}
		\caption{Lane-change (HTD)}
        \label{fig:htd}
    \end{subfigure}
    \caption{\textbf{interPlan scenario types.} The ego vehicle and its navigation route are shown in \ellisorange{orange} and \ellispurple{purple}, surrounding vehicles and pedestrians are \ellisblue{blue} and \ellisgreen{green} respectively.  Traffic cones and stop lines are depicted in \ellisred{red}.}
	\label{fig:interPlan_scenarios}
    \vspace{-0.5cm}
\end{figure*}
\subsection{Problem Setup}
Generalizing to difficult unseen scenarios is a crucial capability to enable real-world autonomy.
However, extracting useful scenarios for benchmarking from real-world recordings is practically infeasible due to their low probability.
Staging them explicitly in real contexts is costly and often morally questionable.
At the same time, generating such scenarios from scratch raises the question of realism.
Therefore, we re-use scenarios from a large-scale real-world dataset and augment them to create difficult and rare scenarios that represent the long tails of the distribution of real-world situations, such as encountering construction zones, jaywalking pedestrians, and accident sites.
We chose the large-scale \nuplan dataset~\cite{caesar2021nuplan} and the respective closed-loop simulator with reactive agents.

Being able to make plans in anticipation of surrounding agents' intentions and reactions is a fundamental pillar of real-world driving and essential to solving scenarios like unprotected turns, overtakes, or lane changes~\cite{rhinehart2021contingencies}.
Therefore, we also evaluate state-of-the-art planning methods in highly interactive scenarios, such as highway merges at different traffic densities or overtaking parked vehicles.
In the following, we describe how we augment the \nuplan scenarios to generate challenging interactive and rare scenarios to build our \interplan benchmark.

\subsection{Scenario Generation}
\boldparagraph{Realistic long-tail scenarios}
In our benchmark, we aim to test long-tail scenarios, which are underrepresented in real-world driving and thus might not occur in large-scale training data.
Such scenarios include encountering a lane blocked by construction zones (cf.\ Fig.~\ref{fig:constr}), nudging around (cf.\ Fig.~\ref{fig:nudge}) or overtaking (cf.\ Fig.~\ref{fig:overtake}) a parked vehicle, and facing jaywalkers at bus stops (cf.\ Fig.~\ref{fig:jayw}).
We generate these scenarios for our benchmark by adding objects such as traffic cones, parked vehicles, or stopped busses to the original \nuplan scenarios while leaving the EV state unchanged.
Moreover, pedestrians who start crossing the street right when the EV approaches test the planner's ability to anticipate such non-compliant behaviors.
In addition, we include scenarios where the planner has to navigate around an accident site (cf.\ Fig.~\ref{fig:crash}).
This is particularly challenging because crashed vehicles with intersecting bounding boxes are found rarely, if at all, in real-world datasets.
Thus, sophisticated reasoning is required to understand that these vehicles have collided and cannot be expected to start moving, so they must be overtaken carefully.
We design the crash sites in such a way that they represent common accidents such as rear-end collisions or collisions with crossing traffic.

\boldparagraph{Interactive lane-change scenarios}
A major shortcoming of the original \nuplan benchmark is that it can be solved almost perfectly by simple lane-following, thus with little interaction with surrounding traffic.
For instance, the challenge-winning method PDM-Closed~\cite{Dauner2023CORL} is by design unable to perform lane changes but achieves a nearly perfect score.
To counteract this, we incentivize lane changes by augmenting the goal of original \nuplan scenarios.
For instance, in the scenario shown in Fig.~\ref{fig:ltd}, the EV has to do 3 lane changes.
Moreover, we initialize the scenarios with different levels of traffic densities, i.e., low (LTD), medium (MTD), and high traffic density (HTD), by spawning traffic agents around the EV. Agents are randomly placed along each lane with sufficient spacing to avoid collisions, with maximum distances of 33m, 50m, and 100m for HTD, MTD, and LTD.

\boldparagraph{Augmented traffic agent behavior}
\label{sec:agent_policies}
Real-world drivers exhibit diverse behaviors ranging from conservative and careful driving to assertive and reckless maneuvers.
Planning methods have to be able to handle these varying conditions, especially without knowing what they are about to encounter.
Our benchmark tests the ability of planning methods to make safe decisions under these varying conditions by applying different policies to control the surrounding traffic agents.
Our primary objective is not realistic traffic simulation, which is an ongoing research field itself, but to create diverse and reasonable traffic agents that can effectively test a planner's interactive behavior.
Hence, in addition to \nuplan's default reactive IDM policy, which breaks as soon as the EV enters its lane, we implement an assertive policy, which only reacts to the EV when it has fully merged into the lane.
Simulation of merging scenarios can be carried out with all agents being controlled by the conservative reactive policy, the assertive policy, or a mixed policy, which randomly assigns a policy to each agent.
The lane-change scenarios in our benchmark combine these traffic agent policies with different traffic densities.
Thus, lane changes in HTD require complex interaction and potentially even prioritizing safety over progress.
At the same time, LTD scenarios test a planner's general ability to merge into another lane.

\subsection{The interPlan Benchmark} 
We test an exhaustive list of \sota planners, including learning-based (PDM-Open~\cite{Dauner2023CORL}, UrbanDriver~\cite{Scheel2021CORL}, GC-PGP~\cite{Hallgarten2022ITSC}, Gameformer~\cite{huang2023gameformer}, DTPP~\cite{Huang2023ICRA}) and rule-based planners (IDM~\cite{treiber2000congested}, IDM+MOBIL~\cite{kesting2007general}, PDM-Closed~\cite{Dauner2023CORL}).
While some strictly follow lane-centerlines, others are not limited regarding their planning space.
We intend to evaluate how well these planners can generalize to unseen scenarios, which is an indispensable capability for safe deployment.
As a consequence, we intentionally test all planners in a zero-shot setting, i.e., they are trained on the large-scale real-world \nuplan train set without fine-tuning on the augmented benchmark scenarios.
For the \interplan benchmark, we select 10 scenarios for each of the following types: Avoiding a construction zone (Constr.), encountering an accident (Acc.), jaywalking pedestrian (Jayw.), nudging around a parked vehicle (Nudge), overtaking a parked vehicle through the oncoming lane (Overt.), and lane changes in low (LTD), medium (MTD) and high (HTD) traffic density.
Among the 10 LTD, MTD, and HTD scenarios, there are 3, 3, and 4 with conservative agents, assertive agents, and mixed agents, respectively.

\subsection{Metrics}
The \interplan benchmark builds upon the established \nuplan metrics, which comprise compliance with driving direction and drivable area, speed limit, as well as safety (collisions, time-to-collision), and comfort (accelerations, jerk).
Metrics for comfortability, speeding, progress, and keeping the TTC within bounds are aggregated into a weighted average.
The other metrics (collisions, compliance with driving direction and area, being stationary for too long) do not contribute to the weighted average, but immediately reduce the score to 0 if performance is below a threshold.
E.g., if the vehicle causes a collision, the entire scenario will be evaluated with a score of 0.
We extend this driving score as follows:
We measure which percentage of lane changes required to get to the goal was completed and include it in the weighted average before applying the penalties.
Moreover, we add an additional multiplicative penalty based on the minimal progress needed to pass the obstacles (parked cars, construction zones, etc.).
Hence, if the vehicle gets stuck in front of an obstacle, the scenario is evaluated with a score of 0\%.
Finally, we deactivate the driving direction compliance penalty for overtakes (Overt.) and accident sites (Acc.), where the vehicle has to go in an oncoming lane to pass the obstacle.

\subsection{Comparison to Val14 Mining}
\begin{table}[t]
    \centering
    \resizebox{0.9\columnwidth}{!}{
    \begin{tabular}{l c c}
        \toprule
         Scenario type & Val14 mining & interPlan (ours)\\
         \midrule
         Total scenarios & 51 & 80 \\
         Noisy objects & 7 & 0 \\
         Unsafe initialization & 13 & 0 \\
         Unsafe IDM behavior & 8 & 0 \\
         Basic driving scenarios & 13 & 0 \\
         Jaywalkers (non-react./react.) & 6 / 0 & 0 / 10 \\
         Non-reactive merge & 4 & 0 \\
         Interactive lane change & 0 & 30 \\
         Nudge & 0 & 10 \\
         Overtake & 0 & 10 \\
         Construction site & 0 & 10 \\
         Accident site & 0 & 10 \\
         \bottomrule
    \end{tabular}
    }
    \caption{Comparison to Val14 mining}
    \label{tab:mining}
    \vspace{-0.5cm}
\end{table}
A straightforward alternative to our scenario construction method is mining a large dataset for challenging scenarios~\cite{cheng2023rethinking}.
We mine the Val14 test split with the \sota PDM-Closed planner and identify 51 scenarios where the score is below 60\%.
See Tab.~\ref{tab:mining} for a comparison between the mined scenarios and \interplan.
\newline
20 of these scenarios turn out to be simulation failures, with initialization of the ego vehicle too close to an obstacle or even off-road (13) or where the sudden appearance of an object leads to an imminent collision (7).
Another eight scenarios contain unrealistically unsafe behavior of the IDM agents, such as running a red light.
\newline
In 13 of the remaining 23 scenarios, the low performance is merely caused by basic planner failures, such as getting stuck at a stop line or slightly going offroad at a tight turn. Thus, the scenarios are not challenging or of particular interest.
Finally, collisions with pedestrians appear in six scenarios, and collisions during merges in four.
In \nuplan simulations, pedestrians always non-reactively follow their recorded logs, which can lead to curious collisions if the controlled vehicle does not exactly follow the recording vehicle's path.
Conversely, in our jaywalker scenarios, the movement of pedestrians is triggered by the approaching EV to always allow a reaction, and the locations are selected in such a way that jaywalkers can be anticipated (e.g., \ at bus stops).
Finally, the four collisions at merges are related to a weakness of the IDM policy used to simulate surrounding agents:
They do not react to vehicles in the neighboring lane, not even if the lanes are about to merge.
Our lane-change scenarios force the EV to enter the lane of a neighboring IDM agent.
By entering the neighboring lane, the EV becomes visible to the IDM so that it is able to react to it.
Because different agent policies are employed (cf.Sec.~\ref{sec:agent_policies}), this creates interactive scenarios where the EV must show diverse strategies to succeed, such as nudging into the lane and waiting for the neighboring vehicle to break and create a gap.
\newline
Not having found a single interactive lane change, we extend the search to all scenarios tagged with lane changes where PDM-Closed achieves a score below 90\%.
In all cases, the EV either gets stuck in a pickup/dropoff zone, or the bad score stems from uncomfortable driving or slow progress.
Although the expert does change lanes, this is not required to follow the route.
In contrast, \interplan's LTD, MTD, and HTD scenarios enforce lane changes via the navigation route and explicitly reward them in the score.
\begin{table*}[t!]
\centering
\resizebox{\textwidth}{!}{
    \large
    \begin{tabular}{l|r|l>{\columncolor[gray]{0.83}}l|lllll||lll|lll}
        \toprule
        & \textbf{Method} & \textbf{Val14} $\uparrow$ & \textbf{interPlan} $\uparrow$ & Constr. $\uparrow$ & Acc. $\uparrow$ & Jayw. $\uparrow$ & Nudge $\uparrow$ & Overt. $\uparrow$ & LTD $\uparrow$ & MTD $\uparrow$ & HTD $\uparrow$ & Driv. $\uparrow$ & Goal $\uparrow$ & No-Col. $\uparrow$\\ 
        \midrule
        \parbox[t]{2mm}{\multirow{5}{*}{\rotatebox[origin=c]{90}{learned}}}
        & \texttt{Urban Driver}~\cite{Scheel2021CORL} & 50 & \xspace 4 & \xspace 0 & \xspace 0 & \xspace 0 & \xspace 0 & \xspace 0 & \xspace 0 & 29 & \xspace 0 & \xspace 30 & \xspace 30 & \xspace 47\\
        & \texttt{GC-PGP}~\cite{Hallgarten2022ITSC} & 55 & 10 & \xspace 0 & \xspace 0 & \xspace 0 & \xspace 0 & \xspace 0 & 18 & 16 & 44 & \xspace 73 & \xspace 17 & \xspace 67\\
        & \texttt{GameFormer}~\cite{huang2023gameformer} & 75 & 11 & \xspace 0 & \xspace 0 & 48 & \xspace 0 & \xspace 0 & \xspace 0 & 20 & 21 & \xspace 30 & \xspace 30 & \xspace 87\\
        & \texttt{PDM-Open}~\cite{Dauner2023CORL} & 54 & 25 & 13 & \xspace 0 & 56 & 36 & \xspace 8 & 29 & 29 & 26 & \xspace 60 & \xspace 40 & \xspace 43\\
        & \texttt{DTPP}~\cite{Huang2023ICRA} & 73 & 25 & 18 & 18 & 44 & 10 & \xspace 0 & 40 & 36 & 34 & \xspace 60 & \xspace 33 & \xspace 93\\
        \midrule
        \parbox[t]{2mm}{\multirow{3}{*}{\rotatebox[origin=c]{90}{rule}}}
        & \texttt{IDM}~\cite{treiber2000congested} & 77 & 31 & \xspace 0 & \xspace 0 & 66 & \xspace 0 & \xspace 0 & 61 & 61 & 61 & 100 & \xspace \xspace 0 & 100\\
        & \texttt{IDM+MOBIL}~\cite{kesting2007general} & 75 & 31 & 21 & \xspace 0 & 66 & \xspace 0 & \xspace 0 & 71 & 21 & 70 & \xspace 93 & \xspace 52 & \xspace 80\\
        & \texttt{PDM-Closed}~\cite{Dauner2023CORL} & 92 & 42 & 18 & \xspace 0 & 48 & 74 & \xspace 9 & 62 & 62 & 62 & 100 & \xspace \xspace 0 & 100\\
        \bottomrule
    \end{tabular}
}
\vspace{0.2cm}
\caption{\textbf{The interPlan benchmark.} Val14 denotes the score reported on common scenarios for reference. 
Besides the aggregated interPlan score, we also report the scores on each scenario type. The sub-scores drivable area compliance (Driv.), reaching the goal lane (Goal), and collision avoidance (No-Col.) refer only to the lane-change scenarios (LTD, MTD, HTD).}
\label{tab:sota}
\vspace{-0.5cm}
\end{table*}
\section{Results}
\subsection{Generalization of \sota Planning Methods}
Our experimental results are shown in Tab.~\ref{tab:sota}.
We find that PDM-Closed achieves the best results with an overall score of 42\%.
However, in contrast to previous results on the Val14 test-split and the \nuplan leaderboard, it is far from achieving a perfect score.
In particular, it achieves the best score among all \sota planners in the Construction Zone, Overtake, and Nudge scenarios.
This is mainly due to its ability to sample different lateral offsets from the centerline it is following, enabling it to pass some of the obstacles in the lane.
Moreover, it outperforms all other methods in the LTD, MTD, and HTD lane change scenarios.
However, due to its limitation of not being able to do lane changes, this merely reflects its ability to find an optimal trajectory concerning comfort and safety since it is unable to reach the goal lane in these scenarios.
In addition, almost all methods in our benchmark fail in the Acc.\ and Overt.\ scenarios, resulting in a score of 0\%.
Exceptions are DTPP for Acc.\ and PDM-Open and PDM-Closed for the Overt.\ scenarios.
While DTPP successfully makes a lane change before the stationary vehicles in the Acc.\ scenario shown in Fig.~\ref{fig:dtpp_acc}, PDM-Closed stops before them (see ~\ref{fig:pdm_acc}).
Similarly, PDM-Open and PDM-Closed pass the parked car in the easiest of the 10 overtake scenarios, i.e., the one with no oncoming traffic.
For PDM-Open, this is not intentional but instead caused by a failure to follow the curved centerline since this model does not consider surrounding vehicles.
Surprisingly, PDM-Closed performs worse than IDM in the scenarios with Jaywalkers even though it employs an IDM policy for centerline following.
In contrast to IDM, PDM uses a future horizon of only 2.0s to evaluate potential longitudinal profiles.
In two scenarios, a collision is unavoidable when the pedestrian is inside this horizon.
In a further scenario, PDM-Closed avoids a collision by evading the pedestrians at high speed (see Fig.~\ref{fig:pdm_ped} instead of stopping and waiting for them to cross (see Fig.~\ref{fig:gameformer_ped}).
Overall, our results reveal critical shortcomings in the generalization capabilities of \sota planning methods.

\subsection{Interactive Lane Changes}
\begin{figure*}[t!]
	\centering
    \begin{subfigure}[t]{\quarterimgwidth}
		\includegraphics[width=\textwidth]{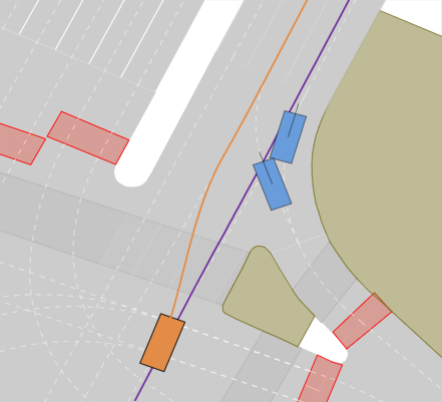}
		\caption{\footnotesize DTTP Acc.}
        \label{fig:dtpp_acc}
    \end{subfigure}\hfill\begin{subfigure}[t]{\quarterimgwidth}
		\includegraphics[width=\textwidth]{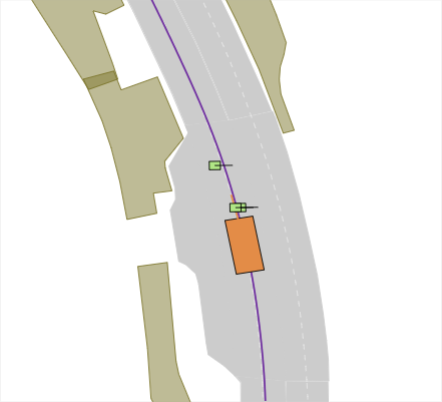}
		\caption{\footnotesize GameFormer Jaywalker}
        \label{fig:gameformer_ped}
    \end{subfigure}\hfill\begin{subfigure}[t]{\quarterimgwidth}
        \includegraphics[width=\textwidth]{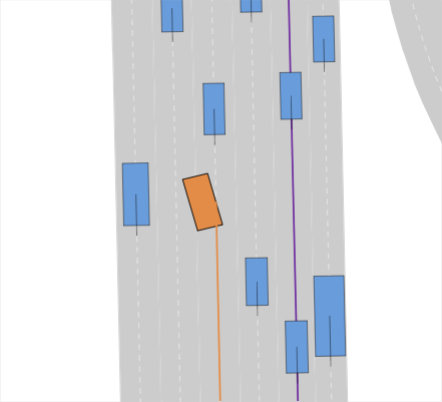}
		\caption{\footnotesize conservative traffic}
        \label{fig:conservative_traffic}
    \end{subfigure}\hfill\begin{subfigure}[t]{\quarterimgwidth}
		\includegraphics[width=\textwidth]{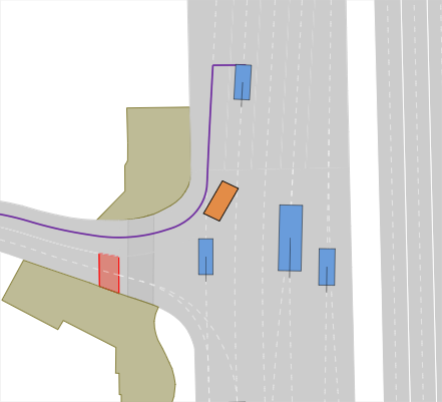}
		\caption{\footnotesize GC-PGP lane change}
        \label{fig:gc_pgp_lane_change}
    \end{subfigure}
    \newline
    \begin{subfigure}[t]{\quarterimgwidth}
		\includegraphics[width=\textwidth]{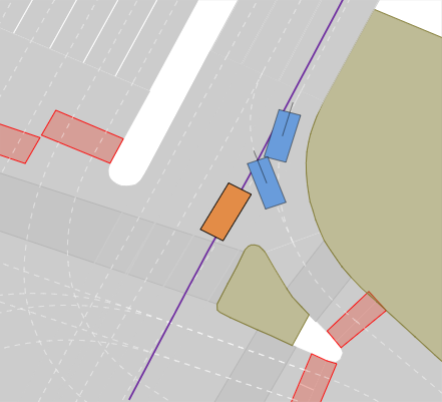}
		\caption{\footnotesize PDM-Closed Acc.}
        \label{fig:pdm_acc}
    \end{subfigure}\hfill\begin{subfigure}[t]{\quarterimgwidth}
		\includegraphics[width=\textwidth]{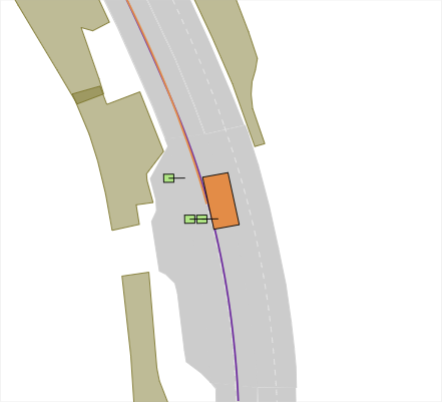}
		\caption{\footnotesize PDM-Closed Jaywalker}
        \label{fig:pdm_ped} 
     \end{subfigure}\hfill\begin{subfigure}[t]{\quarterimgwidth}
		\includegraphics[width=\textwidth]{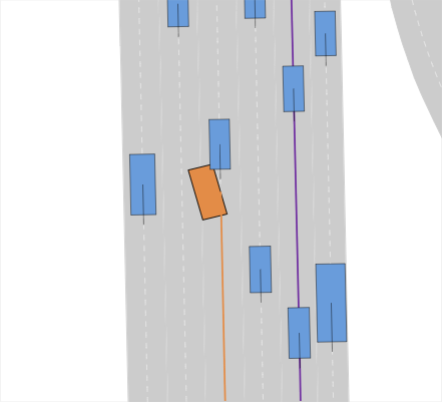}
		\caption{\footnotesize assertive traffic}
        \label{fig:assertive_traffic}
    \end{subfigure}\hfill\begin{subfigure}[t]{\quarterimgwidth}
        \includegraphics[width=\textwidth]{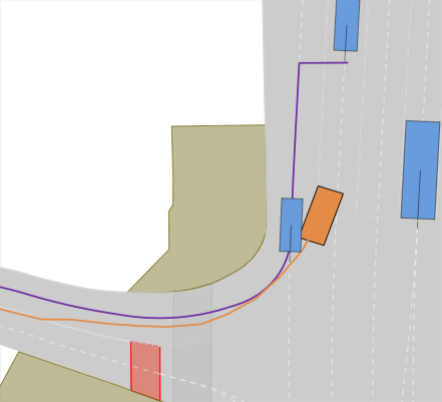}
		\caption{\footnotesize UrbanDriver lane change}
        \label{fig:ud_lane_change}
    \end{subfigure}
    \caption{\textbf{Qualitative results.} 
    Fig.~(a), (b), (c) show successfully avoiding an accident site, waiting for jaywalkers, and executing a lane change surrounded by conservative agents.
    Failure cases are stopping before the accident site (e), narrowly avoiding the pedestrians (f), and colliding with an assertive vehicle approaching quickly from behind (g). Fig.~(d) and (h) show failures of learning-based planners, such as suddenly stopping (d) or causing a collision (h).
    The planned trajectory is depicted in \ellisorange{orange}, the navigation route in \ellispurple{purple}.}
	\label{fig:qual_res}
    \vspace{-0.5cm}
\end{figure*}

In the lane change scenarios (i.e., LTD, MTD, HTD), we often observe trajectories that result in collisions or violations of the drivable area for the learning-based planners UrbanDriver, GameFormer, and PDM-Open (see Fig.~\ref{fig:ud_lane_change}). GC-PGP often stops and gets stuck (see Fig.~\ref{fig:gc_pgp_lane_change}).
On the other hand, rule-based centerline planners (PDM-Closed, IDM) do not cause these infractions and thus incur smaller penalties.
However, they do not reach the goal lane even once.
MOBIL extends IDM with a criterion to achieve safe lane changes, depending on the surroundings.
We use a simple implementation that, given that MOBIL approves the lane change, tasks the controller with reaching waypoints on the neighboring lane.
It achieves the highest rate of reaching the goal lane at 52\%.
However, this comes at the cost of some drivable area infractions and collisions.
We find that some collisions are explained by the downstream controller not being able to merge to the neighboring lane without oscillating, and others are caused by erroneously expecting vehicles approaching from behind to break (see Fig.~\ref{fig:assertive_traffic}).
Nonetheless, it has the highest rate of reaching the goal lane and performs well if the surrounding vehicles act conservatively (see Fig.~\ref{fig:conservative_traffic}).
We conclude that most \sota planners lack sophisticated lane change trajectory planning, such as synchronizing to a gap in preparation, proactively influencing other vehicles' behavior, and aborting a lane change when necessary.

\subsection{Rule-Based vs.\ Learning-Based Planning}
Notably, the rule-based planners IDM, IDM+MOBIL, and PDM-Closed outperform the learning-based contestants.
This is especially surprising for IDM and PDM-Closed, since they are by design limited to simple lane-following without lane changes or overtakes and thus unable to solve scenarios where a construction zone, a parked vehicle, or an accident blocks the current lane.
Moreover, the lowest scores are achieved by planners that regress waypoints (GC-PGP, UrbanDriver) instead of selecting a trajectory among a set of feasible samples.
This indicates that they have the lowest generalization capabilities and are susceptible even to small distributional shifts.
The best learning-based planner is DTPP, which uses a learned cost function alongside trajectories generated by a sampler.
Overall, none of the models demonstrates sufficient world understanding to navigate successfully through these difficult scenarios.

\section{LLM-Based Planning}
In the previous section, we assessed state-of-the-art planning methods under difficult conditions and revealed that there is a critical lack of generalization ability to solve rare and difficult scenarios.
With the rise of foundation models and the tremendous interest surrounding them recently, exploiting their impressive world understanding and generalization capabilities in the field of autonomous driving became a vital field of research.
As the scenarios we proposed in our benchmark demand a thorough understanding of traffic scenarios and complex reasoning (such as anticipating a pedestrian to unlawfully cross the street at a bus stop), approaching the planning task with foundation models appears to be a promising strategy.
Prior work achieves impressive results using LLMs for open-loop trajectory planning in real-world driving scenarios.
However, the current state of the art on regular \nuplan scenarios suggests that such common scenarios can be solved efficiently with simple rule-based planners, and the promise of better generalization to long-tail scenarios is yet to be put to test.
In the following, we explore the capabilities of LLMs in the context of closed-loop behavior planning in such scenarios.
First, we establish a simple baseline, which is an adaptation of the open-loop GPT-Driver~\cite{mao2023gpt} planner.
Then, we describe our LLM-based behavior planning method.
We evaluate both methods on the same \interplan benchmark we used for the \sota planning methods.
By releasing our code, we hope to facilitate further research on improved LLM-based planning for automated driving.

\subsection{LLM Waypoints Planning}
GPTDriver~\cite{mao2023gpt} fine-tunes and instructs an LLM to predict a set of future waypoints.
The prompt comprises general task instruction, perception context, and ego-states.
Chain-of-thought reasoning is used to improve the interpretability and quality of the output.
We apply the method with the same prompt after applying minor changes to adapt it to the \nuplan requirements, i.e., we change the output to an 8-second trajectory at 2 Hz and include route information into the prompt.
Thus, we provide the planner with a `Mission Goal` describing which lanes are on route.
We test various LLMs, including Llama-7B, Llama-13B~\cite{touvron2023llama}, and GPT-3.5.
We call this baseline \texttt{LLMWaypointsPlanner} and fine-tune it on a set of 600 diverse and unaugmented \nuplan scenarios for 3 epochs.
For Llama we employ QLoRa~\cite{dettmers2024qlora} using a rank of 64 and a learning rate of 2e-4.
For GPT, we use openAI's fine-tuning API.

\subsection{LLM Behavior Planning}
In addition, we propose a two-stage hybrid planner that combines an LLM behavior planner with a rule-based motion planner.
In the first stage, the traffic scenario is described in a natural language prompt.
As in the case of our \texttt{LLMWaypointsPlanner} baseline, it comprises general task instruction, perception context, and ego states, as well as route information.
Finally, the model is instructed to select a suitable behavior from a list of available behaviors.
We define the following behaviors: follow the current lane, merge left, merge right, overtake obstacle, stop and wait.
We filter the list based on the availability of neighboring lanes and the presence of obstacles in the current lane before presenting the options to the LLM.
Each option is associated with a centerline and an offset from it.
For instance, `overtake obstacle` refers to the current centerline and the offset required to pass the obstacle, whereas `merge left` refers to the left neighboring centerline with no offset.
The second stage applies a rule-based motion planner to find an optimal trajectory conditioned on the selected behavior parameters.
We leverage PDM-Closed~\cite{Dauner2023CORL}, the top-performing method in our benchmark, for motion planning.
This algorithm samples lateral offsets and longitudinal speed profiles locally around the chosen behavior and selects a trajectory based on a cost function.
The motion planner runs at 10 Hz, while the LLM behavior planner is queried at 1 Hz.
We employ the LLM in a zero-shot setting, as we want to test the generalization of planning methods to unseen difficult scenarios.
Moreover, labeling the regular \nuplan scenarios would almost exclusively result in `follow current lane` samples.


\begin{table*}[t!]
\centering
\resizebox{\textwidth}{!}{
    \begin{tabular}{r|l>{\columncolor[gray]{0.83}}l|lllll||lll|lll}
        \toprule
        \textbf{Model} & \textbf{LLM}& \textbf{interPlan} $\uparrow$ & Constr. $\uparrow$ & Acc. $\uparrow$ & Jayw. $\uparrow$ & Nudge $\uparrow$ & Overt. $\uparrow$ & LTD $\uparrow$ & MTD $\uparrow$ & HTD $\uparrow$ & Driv. $\uparrow$ & Goal $\uparrow$ & No-Col. $\uparrow$\\
        \midrule
            \texttt{PDM-Closed}~\cite{Dauner2023CORL} & None & 42 & 18 & \xspace0 & 48 & 74 & \xspace 9 & 62 & 62 & 62 & 100 & \xspace0 & 100\\
        \midrule
            \texttt{HybridLLMPlanner} & LLama-7B & 53 & 27 & 20 & 48 & 93 & 28 & 81 & 48 & 80 & 100 & 43 & \xspace 87\\
            \texttt{HybridLLMPlanner} & LLama-13B & 48 & 36 & 10 & 48 & 82 & 19 & 71 & 40 & 77 & \xspace 97 & 43 & \xspace 83\\
            \texttt{HybridLLMPlanner} & GPT-3.5 & 40 & 52 & \xspace0 & 48 & 25 & \xspace 9 & 76 & 34 & 69 & \xspace 97 & 47 & \xspace 77\\
        \midrule
            \texttt{LLMWaypointsPlanner} & LLama-7B & 16 & \xspace0 & \xspace0 & \xspace0 & \xspace0 & \xspace0 & 29 & 64 & 37 & \xspace 90 & \xspace0 & \xspace 67\\
            \texttt{LLMWaypointsPlanner} & LLama-13B & 17 & \xspace0 & \xspace0 & \xspace0 & \xspace0 & \xspace0 & 31 & 64 & 38 & \xspace 87 & \xspace0 & \xspace 70 \\
            \texttt{LLMWaypointsPlanner} & GPT-3.5 & 22 & \xspace0 & \xspace0 & \xspace0 & \xspace0 & \xspace0 & 64 & 41 & 69 & \xspace 93 & \xspace0 & 100\\
        \bottomrule
    \end{tabular}
}
\vspace{0.2cm}
\caption{\textbf{LLM baselines on the interPlan benchmark.}}
\label{tab:llms}
\vspace{-0.5cm}
\end{table*}
\subsection{Rule-Based Motion Planners can be enhanced with LLMs}
Tab.~\ref{tab:llms} compares the results of both LLM-based planning methods to the best method on our benchmark, i.e., PDM-Closed.
Despite using the same prompt and fine-tuning strategy, the \texttt{LLMWaypointsPlanner} baseline cannot reproduce the strong results from other benchmarks (see~\cite{mao2023gpt}) and trivially outputs a longitudinal profile at a constant heading angle.
Conversely, our \texttt{HybridLLMPlanner} outperforms PDM-Closed, setting a new state-of-the-art on our proposed benchmark.
Surprisingly, the larger 13B Llama model performs worse than the smaller 7B version often making questionable decisions, resulting in collisions or less progress.

\subsection{LLMs lack Traffic Understanding}
While the \texttt{HybridLLMPlanner} outperforms all existing methods, it is merely a strong baseline and unable to solve all the difficult \interplan scenarios.
For instance, we observe that it often toggles between different behaviors, such as overtaking and waiting for oncoming traffic, ultimately getting stuck behind the obstacle.
Moreover, the decision is often inconsistent with the situation analysis, indicating a lack of understanding of traffic.
Thus, we highlight fine-tuning with auxiliary tasks to enhance traffic understanding as an important direction for future research.
\section{Conclusion}
In this work, we presented a novel realistic benchmark that evaluates planning algorithms in highly interactive and rare long-tail scenarios.
We found that existing state-of-the-art methods exhibit critical limitations in their ability to generalize to such scenarios.
Moreover, we provided an LLM-based planning baseline, proposed a novel LLM-based behavior planner, and explored their capabilities.
We hope that our work encourages future research to tackle this generalization problem as we see large potential in the use of multimodal foundation models to better understand traffic scenarios and plan suitable maneuvers, which will be the focus of our future work.

\section{Limitations}
While interPlan extends the widely adopted nuPlan simulator with challenging scenarios, several aspects remain unaddressed. As all scenarios are hand-designed or generated based on heuristics, the benchmark comprises only a limited number of scenarios.
Nonetheless, interPlan is able to reveal critical shortcomings in current state-of-the-art methods. 
Further, interPlan scenarios don't provide ground truth and thus cannot be used for training. Therefore, planners can only be trained on the regular nuPlan scenarios before testing their generalization capabilities in interPlan.
Finally, we believe that data-driven traffic simulation could further increase traffic diversity and make the benchmark more challenging.

\bibliographystyle{IEEEtran}
\bibliography{bibliography}

\end{document}